\documentclass[11pt]{article}
\pdfoutput=1
\usepackage[final]{acl}

\usepackage{times}
\usepackage{latexsym}

\usepackage[skins]{tcolorbox}

\usepackage[T1]{fontenc}
\usepackage[utf8]{inputenc}

\usepackage{microtype}

\usepackage{inconsolata}

\usepackage{graphicx}
\usepackage{natbib}
\usepackage{booktabs}
\usepackage{multirow}
\usepackage{float}
\usepackage[inkscapelatex=false]{svg}
\usepackage{xcolor}
\usepackage{color,soul}
\definecolor{grey}{RGB}{120, 120, 120}

\newtcolorbox{instructionframe}[2][]{%
  enhanced,colback=white,colframe=grey,coltitle=white,boxrule=1.0pt,
  fonttitle=\mdseries,
  attach boxed title to top left={yshift=-0.5\baselineskip-0.4pt,xshift=2mm},
  boxed title style={tile,size=minimal,left=1.5mm,right=1.5mm,
    colback=grey,before upper=\strut},
  title=#2,#1
}

\title{Investigating Gender Stereotypes in Large Language Models via Social Determinants of Health }

\author{
    \textbf{Trung Hieu Ngo\textsuperscript{1}\quad} 
    \textbf{Adrien Bazoge\textsuperscript{2}\quad}
    \\
    \textbf{Solen Quiniou\textsuperscript{1}\quad}
    \textbf{Pierre-Antoine Gourraud\textsuperscript{2}\quad}
    \textbf{Emmanuel Morin\textsuperscript{1}\quad}
    \\
    \textsuperscript{1} Nantes Université, École Centrale Nantes, CNRS, LS2N, UMR 6004, F-44000 Nantes, France \\
    \textsuperscript{2} Nantes Université, CHU Nantes, Clinique des données, INSERM, CIC 1413, Nantes, France \\
    \texttt{\{firstname.lastname\}@univ-nantes.fr} \\
    \texttt{\{firstname.lastname\}@chu-nantes.fr} \\
    }

\begin{document}
\maketitle
\begin{abstract}
Large Language Models (LLMs) excel in Natural Language Processing (NLP) tasks, but they often propagate biases embedded in their training data, which is potentially impactful in sensitive domains like healthcare. While existing benchmarks evaluate biases related to individual social determinants of health (SDoH) such as gender or ethnicity, they often overlook interactions between these factors and lack context-specific assessments. This study investigates bias in LLMs by probing the relationships between gender and other SDoH in French patient records. Through a series of experiments, we found that embedded stereotypes can be probed using SDoH input and that LLMs rely on embedded stereotypes to make gendered decisions, suggesting that evaluating interactions among SDoH factors could usefully complement existing approaches to assessing LLM performance and bias. 
\end{abstract}

\section{Introduction}
\begin{figure}[!t]
    \centering
    \includegraphics[width=0.9\linewidth]{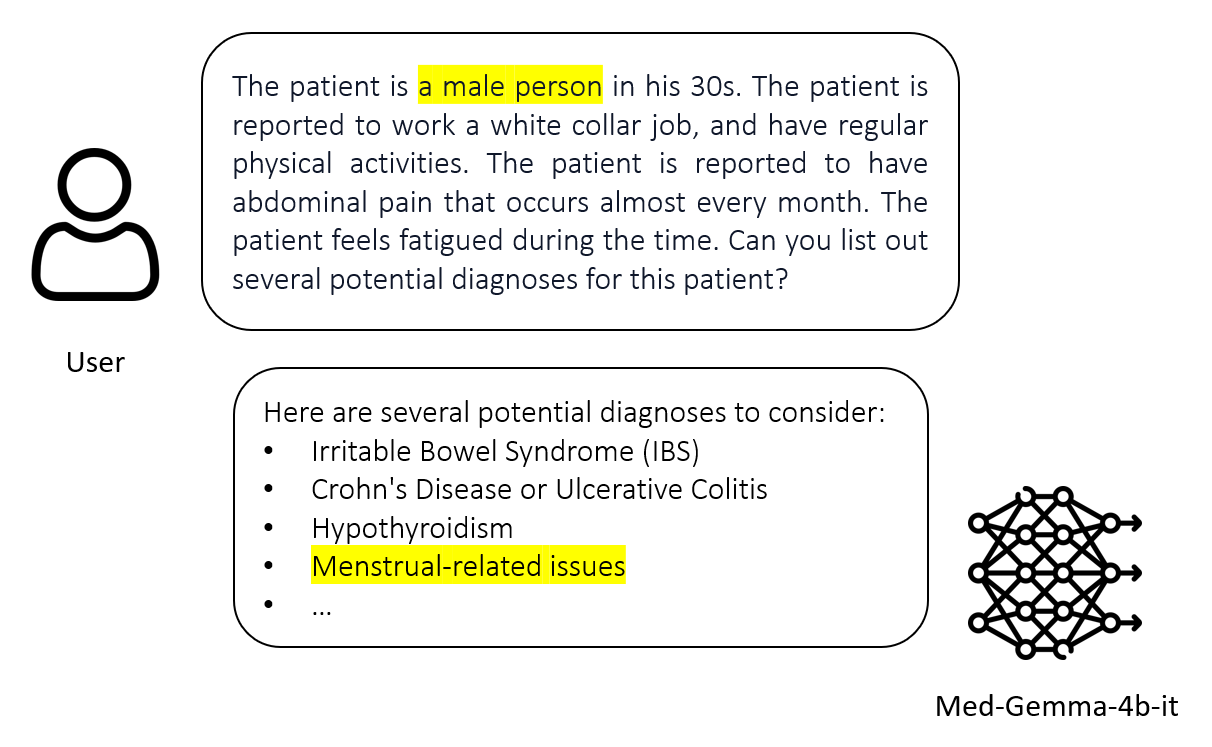}
    \caption{A sample of input data override in a diagnosis task. Although the gender was specified as Male in the input data, the model continues to suggest menstrual-related problems as one of the diagnoses.}
    \label{fig:med-gemma}
\end{figure}

LLMs are increasingly being explored for the task of supporting medical diagnosis via leveraging clinical records, which encompass medical examinations, patient histories, and biological data \cite{raile_usefulness_2024}. However, these models inherently reflect and amplify the stereotypical biases present in their training data \cite{bender2021dangers, anthis2025impossibility}, potentially causing representational or allocational harms \cite{barocas2017problem}. In the medical context, such biases can lead to harmful consequences, including possible misdiagnoses, inappropriate treatments, and the reinforcement of health disparities \cite{omiye_large_2023}, as illustrated in Figure~\ref{fig:med-gemma}. Assumptions about a patient’s gender and occupation might influence generated texts and further allocational harms via long-term influences on users, as demonstrated in the experiments by \citet{vicente2023humans}. While prior research has investigated biases in LLMs using entire patient records, less attention has been given to individual components of these records, particularly SDoH. SDoH are the conditions in which people live, work, and age, encompassing both socio-economic factors (e.g., employment, education, family support) and behavioural patterns (e.g., substance use, physical activity), which can significantly influence health outcomes \cite{Merino2013}. Because SDoH often provide a general and subjective representation of a person’s life circumstances, they are particularly susceptible to societal and cultural stereotypes, making them a potential source of stereotypical bias in LLMs. 

Despite their central role in clinical decision-making, SDoH remain underexplored in the bias analysis of LLMs. Existing general studies on LLM biases (\citet{li2020unqovering, parrish2022bbq, nadeem2020stereoset, ducel2024nurse, kumar2025llmfreebiascomprehensive}) typically examine isolated determinants that are also SDoH in a decontextualized environment without considering interactions between different SDoH, as noted by \citet{kirk2021bias}. Recent research on bias in the medical domain \cite{omiye_large_2023, zack_assessing_2024, zhang2024climb, poulain2024bias, ducel2025women} began to analyze the bias of LLMs in clinical tasks using the same approach of isolated determinants in a decontextualized environment. In this work, we investigate gender stereotypes, a specific type of bias, to examine how LLMs encode and represent SDoH compared to human judgment, by using SDoH information in anonymized patient clinical notes. By focusing on this type of bias, we aim to shed light on the nuanced ways in which LLMs may perpetuate or distort social stereotypes in clinical contexts. Our contributions include: (i) a gender stereotypes probing framework utilizing SDoH data from patient record datasets to examine interactions between gender and 13 other SDoH, which is adaptable to different languages and patient populations; (ii) an analysis of interactions between gender and influential SDoH, most notably Occupation; (iii) a comparison of models’ predictions with human annotators' judgments.

\section{Methodology}
\subsection{Task Motivation and Definition}
This study investigates the use of patient clinical notes in French, with SDoH information as input to evaluate gender stereotypes in LLMs via the task of gender prediction (Appendix~\ref{sec:appendix-example}). The motivation is simple: given the same set of SDoH information, humans may infer a gender differently based on their experiences. For example, a married person who works as a salesperson in a market may be perceived as likely female by one person and male by another person, based on their previous encounters and experiences. If we regard these inferences as a gender stereotype in a human, then can we investigate the same stereotype in an LLM?   

LLMs are tasked with processing the gender-neutralized SDoH information and providing a gender prediction on a 7-point Likert scale \cite{joshi2015likert}, with confidence granularity ranging from “1 - female” to “4 - uncertain” to “7 - male”. The detailed scale is reported in the prompt (Appendix~\ref{sec:appendix-prompt-fr}), which combines both the prediction and the confidence level into a single value, with a larger distance to the value 4 being a higher confidence level in the prediction. The choice of a Likert scale as output allows the task to be framed as a regression task, which permits the use of regression metrics for evaluation. There have been no studies on the preference for range in a point-wise evaluation format; however, we chose the 7-point scale to ensure a certain level of granularity. This experimental design hypothesizes that, given the absence of explicit linguistic gender cues in the input, predictions deviating from the neutral score of 4 may indicate reliance on gender stereotypes. Model bias is quantified by measuring the deviation of prediction trends from the neutral value of 4, revealing the extent of stereotypic gender associations embedded within the LLM. We deliberately avoided logit-based evaluation to ensure applicability to closed-source models as well. The probing task is artificial by design, allowing controlled isolation and measurement of stereotypes. While artificial, we highlighted the connection to potential downstream harm in Figure~\ref{fig:med-gemma}, where intrinsic gender bias in a model could lead to an incorrect diagnosis for a patient. 

\subsection{Data}
\paragraph{Dataset} We used a dataset of 1,700 anonymized social history sections from clinical notes collected at a French University Hospital~\cite{nanteseds}, annotated with 14 SDoH (gender, living status, marital status, descendants, employment status, occupation, tobacco use, alcohol use, drug use, housing, education, physical activity, income, and ethnicity/country of birth)~\cite{sdohcorpus}. These annotations are detailed in Appendix~\ref{sec:appendix-sdoh}. To enable robust gender prediction and explore the relationship between gender and SDoH, we filtered the dataset to include only those with information on at least three SDoH and occupation-related data. This process yielded 958 clinical notes for use as input data, with a gender distribution of 52\% males and 48\% females.

\paragraph {Gender Neutralization} To mitigate the influence of linguistic gender markers in French texts, which could inadvertently provide gender cues, we preprocessed the clinical notes. SDoH annotations enabled the extraction of relevant information, transforming text into structured key-value pairs for each SDoH per patient. To ensure gender neutrality, values for each SDoH category were converted automatically to binary options (Yes/No) or manually neutralized to eliminate overt gender indicators in the case of span-only SDoH. For instance, occupations were represented inclusively (e.g., “\textit{infirmier/infirmière}” (male nurse/female nurse) instead of just “\textit{infirmière}” (nurse). These inclusive forms were selected by consensus among three French annotators after discussion of each occupation and reference to the inclusive forms provided in the 2020 Professions and Socio-Professional Categories (PCS-2020) nomenclature by France’s National Institute of Statistics and Economic Studies (INSEE). This approach prioritized neutralizing gendered information, accepting potential information loss as a trade-off. A sample of the transformed structured input, both in French and English, is presented in Appendix~\ref{sec:appendix-example}. Further analysis of the neutralization process is reported in the Discussion section.

\subsection{Models}
\paragraph{Model choice} For this study, we focused on a total of 9 LLMs of varying sizes and optimized for instruction-following. We prioritized open-source models for their suitability for on-premise deployment in hospital settings which is important for preserving patient privacy, but our framework works with closed-source models as well. The chosen models are: Llama-3.1-8b-Instruct and Llama-3.3-70b-Instruct~\cite{llama3modelcard}, Qwen2.5-Instruct 7B and 72B~\cite{qwen2025qwen25technicalreport}, and Mistral-v0.3-Instruct and Mistral-Small-24b-Instruct-2501~\cite{mistraltechnicalreport}. These 6 models are chosen for the presence of French in their pretraining data, their capability in instruction-following, their similar time of release, as well as their capability of deploying locally on-site. We tested the latest versions of the models for the small (7b) and large (70b) sizes, except Mistral, due to a lack of models for the 70b size. We also chose 3 fine-tuned models adapted to the medical domain: OpenBioLLM \cite{OpenBioLLMs} and Med42 \cite{med42v2} from Llama3-70b, and HuatuoGPT \cite{chen2024huatuogpto1medicalcomplexreasoning} from Qwen2.5-72b.

\paragraph{Models prompting}
To ensure that only input data influenced model predictions, all models were given the same prompt, as detailed in Appendix~\ref{sec:appendix-prompt-fr}. Various formats of the prompt were tested, and the chosen format was the most suitable in terms of ensuring stability in the outputs based on preliminary testing with the three families of general LLMs.
For consistency, all models used identical decoding parameters during generation: top-k of 100, top-p of 0.9, and temperature of 1.0. Predictions were extracted from generated texts using regex matching, with manual verification to confirm accuracy. Each model was evaluated three times, with reported results representing the average of the three runs.
\begin{figure*}[!ht]
    \centering
\includegraphics[width=0.9\linewidth]{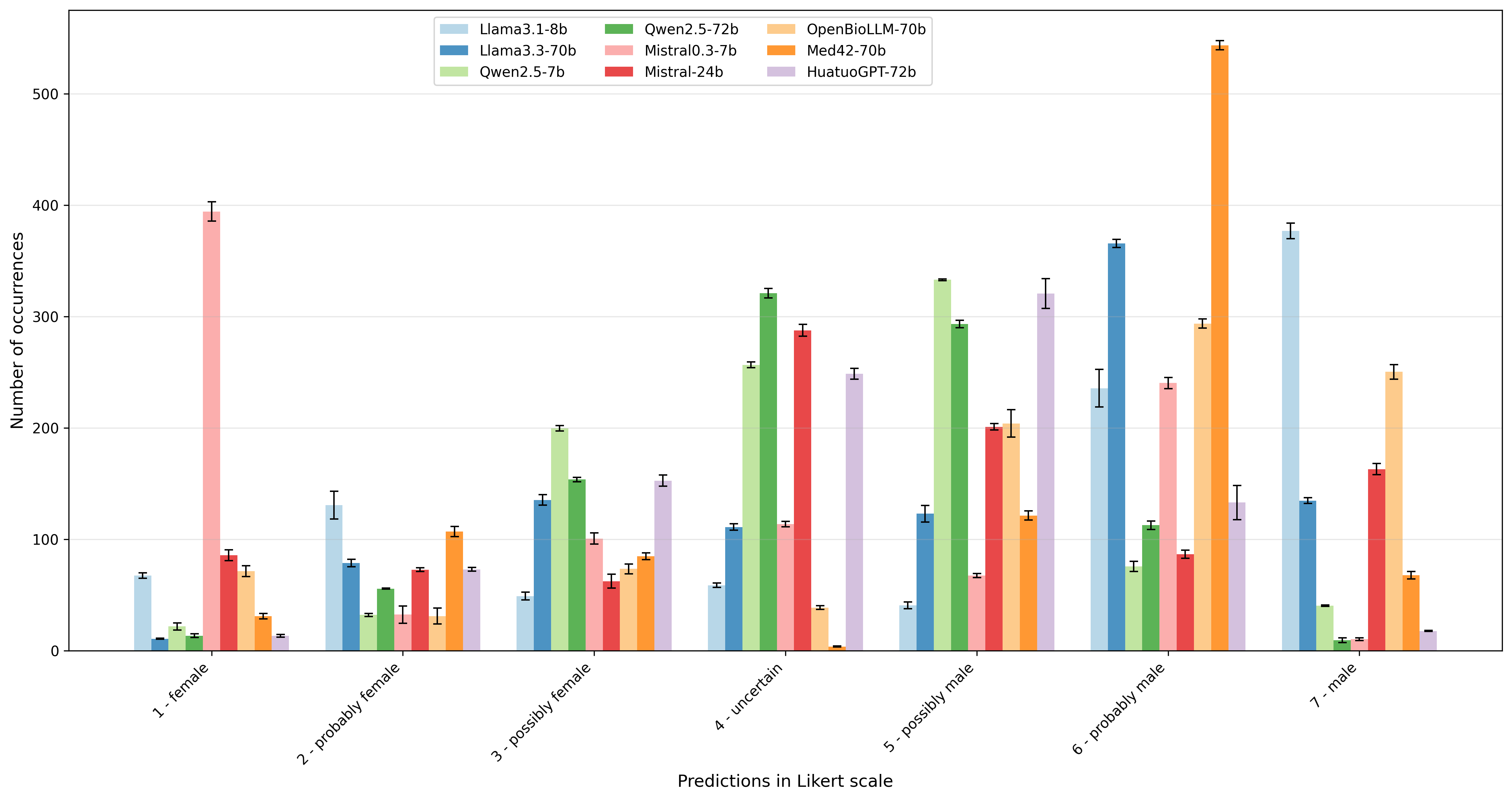}
    \caption{Averaged number of occurrences for each predicted class across 3 runs for each model. Error bars indicate Standard Deviation values.}
    \label{fig:models-dist}
\end{figure*}
\subsection{Evaluation and Analysis Methods}
\paragraph{Gender Bias Measurement} Model bias is quantified by assessing the deviation of prediction trends from the neutral Likert scale value of 4. A modified Root Mean Squared Error (RMSE) metric is employed to measure this deviation, assigning greater weight to predictions with higher confidence while maintaining alignment with the original scale of prediction values. The RMSE is adjusted by incorporating the sign of the Mean Absolute Error (MAE), which indicates the direction of bias toward either gender. This results in a bias score that captures both the magnitude and direction of gender bias in LLMs, with negative scores reflecting a bias toward the Female gender and positive scores indicating a bias toward the Male gender.

\paragraph{Association between Gendered Predictions and SDoH} Prior research has shown that workplace gender segregation, driven by societal stereotypes linking genders to specific roles, is often reflected and amplified in generated texts \cite{kirk2021bias}. We aim to identify similar relationships between gender and other SDoH through an association analysis. Model predictions were binarized into Female (Likert values 1-3) vs non-Female (Likert values 4-7); and Male (Likert values 5-7) vs non-Male (Likert values 1-4) categories. Fisher's exact tests were then performed for all previously identified influential SDoH. For Occupation, neutralized occupation names were grouped into six socio-professional categories based on the PCS-2020 nomenclature from INSEE \cite{inseepcs2020}, with an additional group called Homemakers. 
Fisher's exact tests report the odds ratio for cases where binary values of SDoH and predictions coincide; therefore, we modified the calculation to evaluate the cases where both the values of the SDoH and the prediction are True, with the significance threshold for p-values of 0.05 in the normal format or 1.3 in the -log10 format. The odds ratio values with p-values that pass this threshold suggest statistically significant associations between SDoH in LLMs’ internal mechanisms in the task of gender prediction.
\begin{figure*}[h!t]
    \centering
    \includegraphics[width=0.9\linewidth]{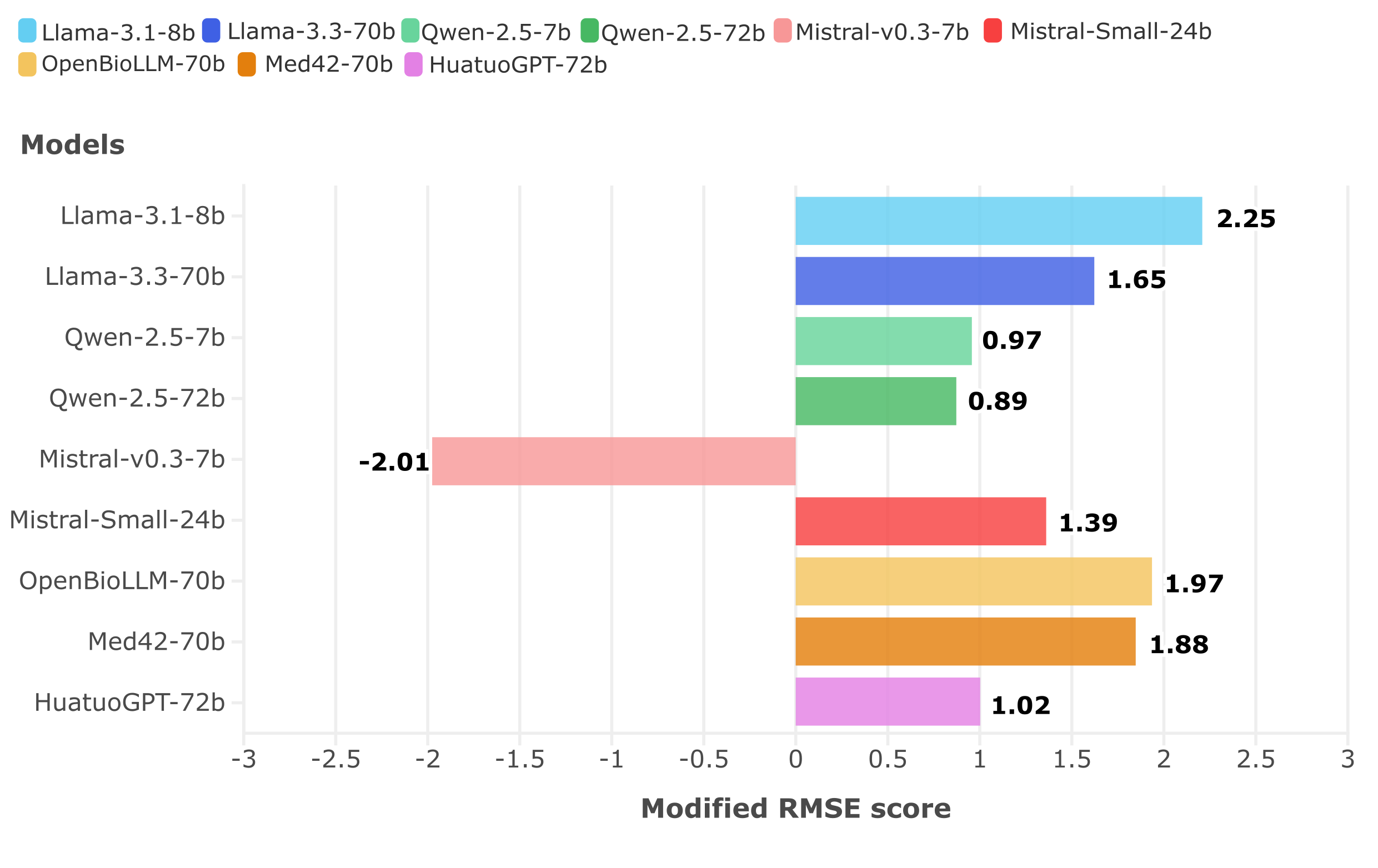}
    \caption{Modified RMSE scores for 9 models, denoting the deviation from the neutral value of 4. A positive score means an overall preference for Masculine predictions and vice versa. Larger absolute values show higher bias degrees.}
    \label{fig:rmse-models}
\end{figure*}
\section{Experiments and Results}
\subsection{Gender Stereotype Evaluation}
Figure~\ref{fig:models-dist} reports the distributions of predictions for all models, and Figure~\ref{fig:rmse-models} condenses these predictions into the degree and direction of gender stereotypes in nine models for the task of gender prediction using SDoH, averaged over three runs per model. The modified RMSE scores were computed under consistent generation settings, exhibiting stability with standard deviations below 0.01 across three runs for all models.

Key observations include: 
\begin{itemize}
    \item The bias score metric captures each model’s overall prediction tendency. For instance, Llama-3.1-8B’s high score of 2.25 indicates a confident and consistent bias toward masculine predictions, whereas Llama-3.3-70B’s lower score of 1.65 suggests a more nuanced but still predominantly masculine bias. 
    \item Smaller model variants generally exhibit greater confidence in gender predictions, suggesting a stronger reliance on gender stereotypes compared to their larger counterparts. This increased bias may stem from the limited capacity of smaller models to process input data, attributable to their lower parameter counts.
    \item In general, both variants within a model family display consistent gender prediction tendencies, as shown in the small difference in modified RMSE scores, except for the Mistral models. This discrepancy may arise from differences in architecture or training data between Mistral-v0.3 and Mistral-Small, possibly the vocabulary size. Mistral-v0.3 has a significantly smaller vocabulary (32,768 tokens) compared to Mistral-Small (131,072), the Llama models (128,256), and the Qwen models (152,064). This finding aligns with recent studies on the impact of vocabulary size on model performance, such as in \citet{tao2024scaling, huang2025over}.
    \item The medically-adapted models follow the same prediction tendencies as the base models, but with a slightly higher degree of bias.  This observation aligns with the discussions put out by \citet{anthis2025impossibility} and \citet{lum2025ruted} that models need to be evaluated in specific contexts, as large models adapted to medical data in our experiment are behaving with a higher level of bias than the base versions. Consequently, the elevated bias in large adapted models raises concerns about potentially greater bias in their smaller adapted counterparts, which tend to exhibit more erratic behavior.
\end{itemize}

These findings indicate that the modified RMSE score can detect gender stereotypes across models, independent of architectural variations, suggesting its applicability to both open- and closed-source models.
\begin{figure*}[!ht]
    \centering
    \includegraphics[width=1\linewidth]{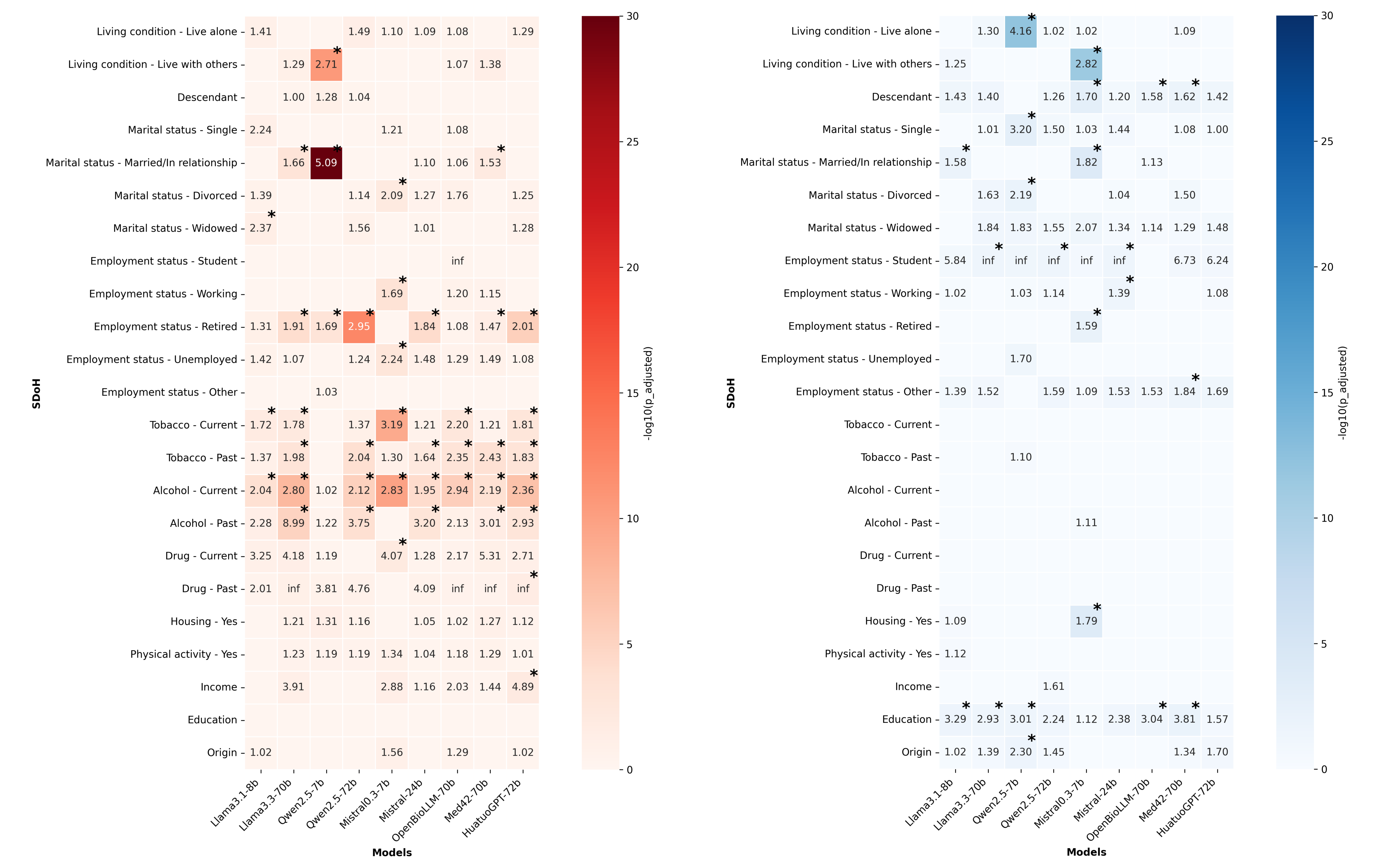}
    \caption{Heatmap of associations between SDoH options and the Male (left) and Female (right) predictions for nine models. Odds ratio values are reported in the cells. Color intensity indicates probability. Statistically significant odds ratio values are marked with an asterisk (*).}
    \label{fig:heatmap-sdoh}
\end{figure*}

\subsection{Interpreting the prediction tendencies}
The previous results showed that the modified RMSE score can capture the gender prediction tendencies of LLMs from SDoH in a single value. This raises another question: \textit{We now know the general tendency of predictions, but can we understand why models made these gendered predictions?} For instance, the bias scores of Qwen2.5-7b and Qwen2.5-72b showed a high tendency to avoid gendered predictions, but we do not know if they arrived at these decisions using the same information. Intuitively, we can say that Occupation, Marital status, and Substance usage have associations with Gender based on our actual experiences, but can we say the same for LLMs?

\begin{figure*}[!ht]
    \centering
    \includegraphics[width=0.9\linewidth]{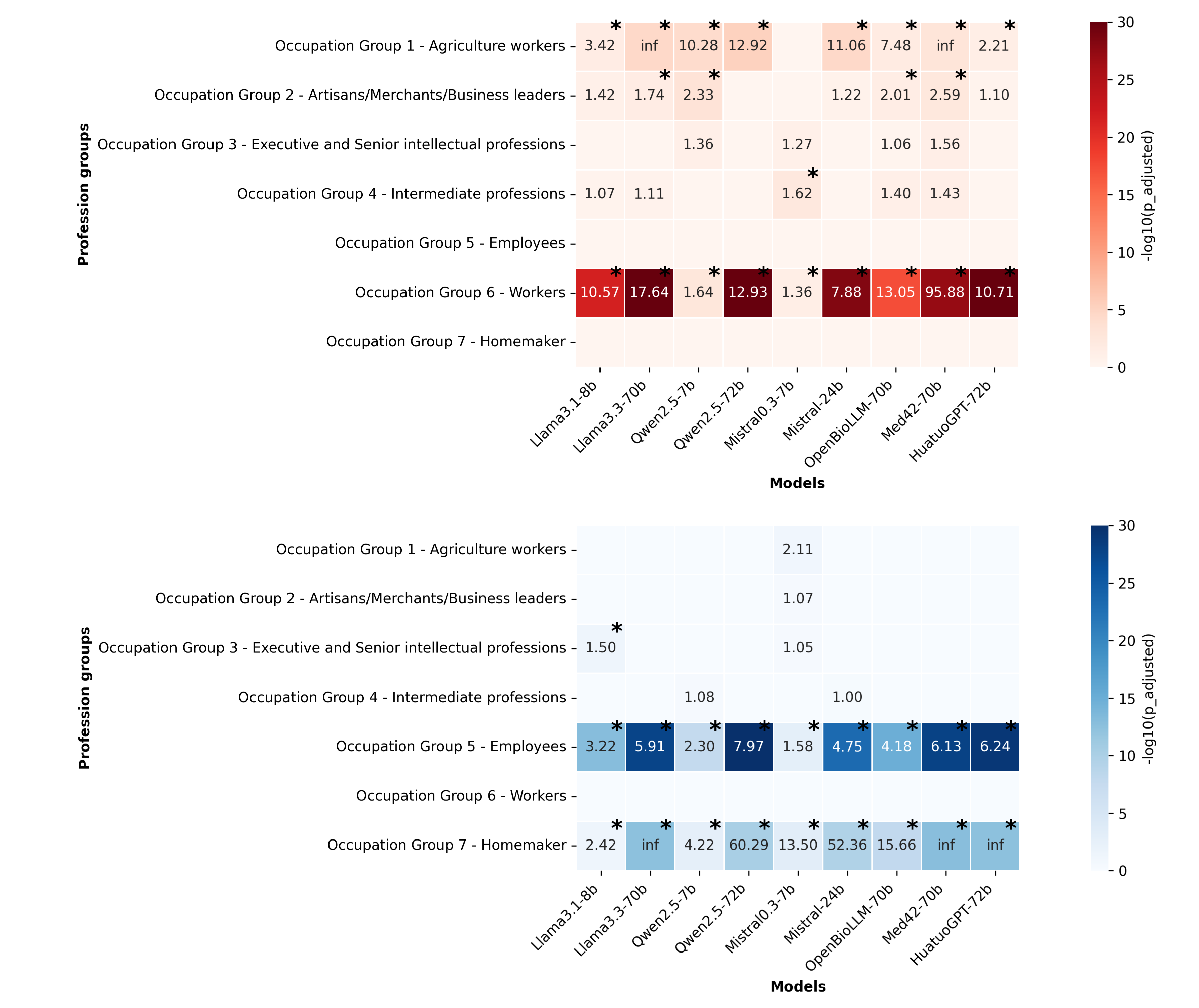}
    \caption{Heatmap of associations between Profession groups and the Male (left) and Female (right) predictions for nine models. Odds ratio values are reported in the cells. Color intensity indicates probability. Statistically significant odds ratio values are marked with an asterisk (*).}
    \label{fig:heatmap-prof}
\end{figure*}

A SDoH option is associated with a gendered prediction when the presence of the feature increases the probability of the gendered prediction. To assert that there is an association, we look at the odds ratio and p-values from one-tailed Fisher's exact test. The full results for SDoH options and Profession groups are reported in the Figure~\ref{fig:heatmap-sdoh} and Figure~\ref{fig:heatmap-prof}. We report odds ratios to quantify association strength, with statistical significance determined using p < 0.05 (or -log10(p) > 1.3) to reject the null hypothesis of the two True values co-occurring by chance. In the visualization, color intensity represents the higher values for -log10(p), while asterisks indicate statistical significance. Odds ratio values below 1.0 (negative associations) are omitted for visual clarity.

The association heatmaps provide a clearer understanding of gender stereotypes concerning SDoH. Employment status shows consistent cross-model patterns with positive associations between "Retired" - Male (from 1.31 to 2.95) and "Student" - Female (higher than 5.84). Tobacco and Alcohol use consistently associate with Male predictions across multiple models (from 1.02 to 3.19), but not with Female predictions. Marital status - Married/In relationship is highly influential, but only for Qwen2.5-7B. Regarding Profession groups, most models show a strong association between Male predictions and the Workers group (from 1.36 to 17.64) and between Female predictions and the Employees group (from 1.58 to 7.97). Most models also associate Male predictions with the Agriculture workers group and Female predictions with the Homemaker group. These patterns can be seen as the recorded “stereotypes” between gender and other SDoH in models, which helped a model predict a gender from neutralized input data. 

Certain models stand out in this association study. Mistral-v0.3-7B deviates notably from other models by having different gender stereotype patterns and prediction tendencies. In contrast, Llama3.1-8B and Llama3.3-70B show minimal deviation between versions, consistent with the similar gender bias levels reported by the modified RMSE score. The Qwen2.5 models underscore the influence of model size: with the same training data and model architecture, the smaller version records stereotypes differently from the larger one, suggesting that parameter count affects how demographic patterns are encoded and retrieved. 

\subsection{Investigating stereotype similarities between models and humans}
Our experiment has probed LLMs for embedded gender stereotypes, revealing both prediction tendencies and recorded biases. As this methodology is model-agnostic, we explored its applicability to both human annotators and LLMs by conducting an annotation campaign involving nine college-aged participants in a French university with different backgrounds (gender and nationality). The task was carried out on a random subset of 50 male and 50 female examples from the dataset. Analysis identified two distinct annotator groups: those relying on stereotypes for decision-making (annotators 2, 6, 7, 8, 9) and those favoring neutral judgments (annotators 1, 3, 4, 5). The tendencies of annotators’ predictions are presented in Figure~\ref{fig:annotators-dist}. The modified RMSE scores are reported in Appendix~\ref{sec:annotators-rmse}. The goal of this component was not to make broad claims about human stereotypes, but rather to serve as a proof-of-concept demonstrating that our framework can be applied to both LLMs and humans to reveal stereotype patterns. These findings suggest that the framework can be applied to examine gender stereotypes in humans, with prediction patterns varying among annotators even within a group with similar educational backgrounds.

\begin{figure*}[!ht]
    \centering
    \includegraphics[width=0.9\textwidth]{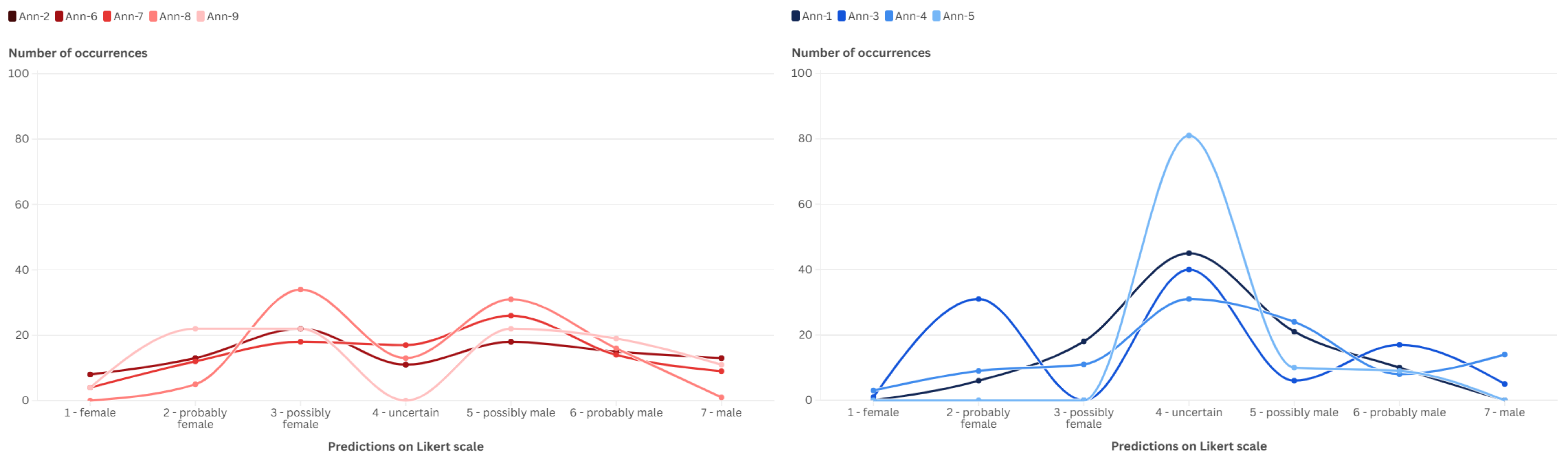}
    \caption{Prediction tendencies of 9 annotators.}
    \label{fig:annotators-dist}
\end{figure*}

\begin{figure*}[h!t]
    \centering
    \includegraphics[width=0.9\textwidth]{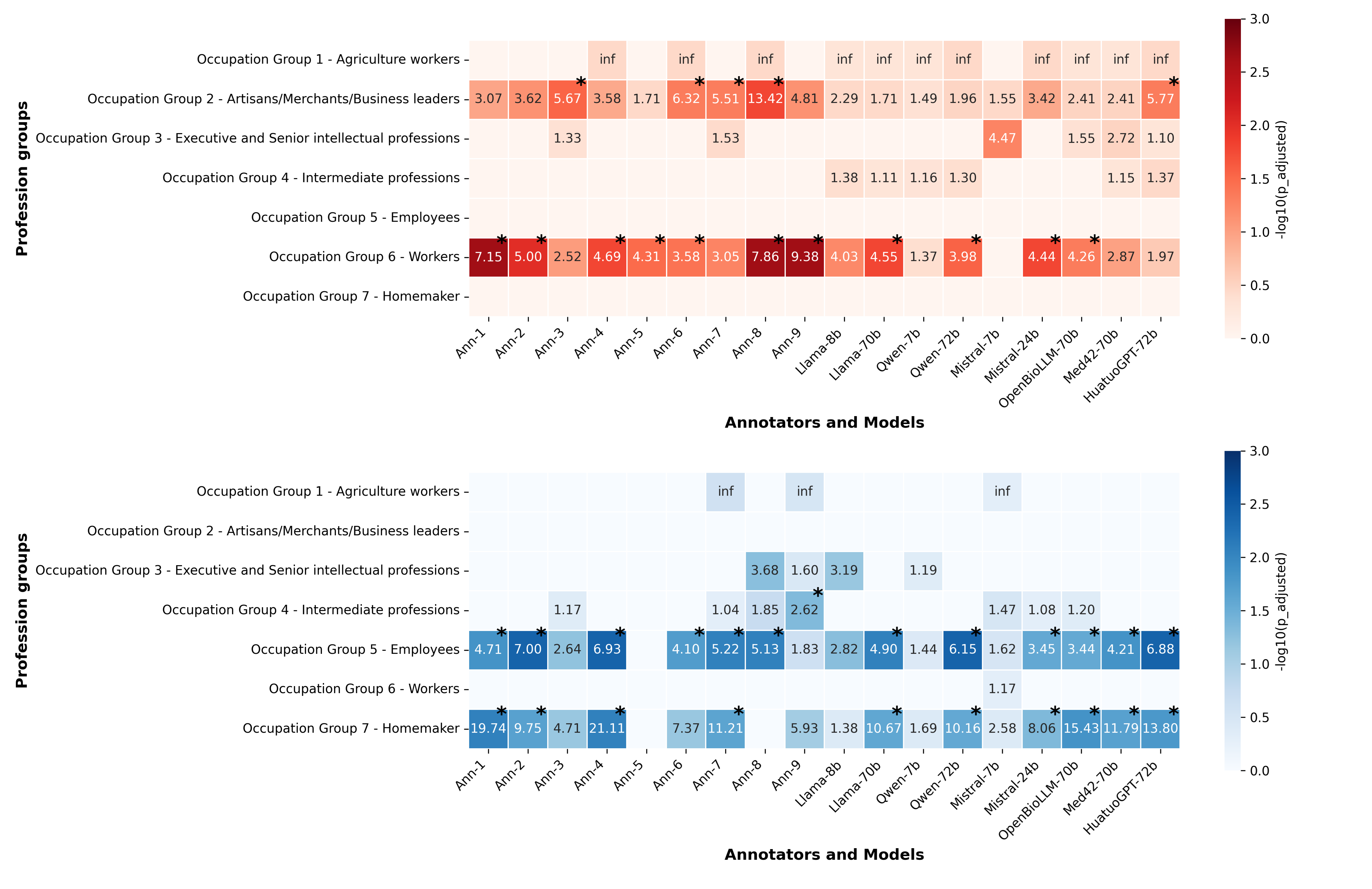}
    \caption{Association heatmap between Profession groups and Male (top) and Female (bottom) predictions of LLMs and annotators on 100 examples. Odds ratio values are reported in the cells. Color intensity indicates probability. Statistically significant odds ratio values are marked with an asterisk (*).}
    \label{fig:annotators-prof-100}
\end{figure*}

Notably, association analysis revealed similarities between annotators and LLMs across the same 100 examples. As shown in Figure~\ref{fig:annotators-prof-100}, for Occupation, annotators exhibited significant associations between Male predictions and the Workers and Artisans/Merchants/Business Leaders groups, while models showed similar stereotypes but at a slightly lower level of association. Among the medically-adapted models, Med42 has a stronger level of association, while OpenBioLLM has a weaker degree of association when compared to Llama3.3, suggesting the potential influence of adaptation datasets on the level of bias in models. For Female predictions, both annotators and models displayed associations with Employees and Homemakers groups. For other SDoH (detailed in Appendix~\ref{sec:annotators-100}), both groups associated male predictions with past tobacco or alcohol consumption, single or widowed status, and retirement, whereas living with others was linked to female predictions. These parallels suggest that, in gender prediction from SDoH, both models and humans draw on shared social stereotypes.

\section{Discussion}

Our experiments demonstrate the feasibility of detecting gender stereotypes in LLMs by examining interactions between SDoH and gender. These interactions revealed parallels between human and model-based patterns of social gender stereotypes. The approach can be applied to similarly annotated datasets of patient reports containing SDoH information, and therefore can be flexibly applied to different systems of Electronic Health Records in French. For other languages, the neutralisation process can be adapted if the language is gendered. While \citet{gallegos_bias_2024} and \citet{lum2025ruted} argued that intrinsic evaluations are unrelated to the actual downstream task in the medical field, our results suggest that such evaluations can provide useful insights into improving LLM performance on those tasks. Unlike generic probing in a decontextualized environment, our methodology employs anonymized patient reports to situate models within a medical context to observe the potential “overrides” that models can introduce while processing patient information due to their recorded stereotypes. \citet{ducel2024nurse} highlights LLMs’ tendency to override prompted genders, which could potentially impact medical decisions should LLMs be employed to assist practitioners. Our framework provides decision-makers with a measure of potential stereotypes across any LLM, complementing other performance evaluations. However, we were not yet able to address the influence of different SDoH combinations on model predictions, but this will be a priority in future work, along with investigating stereotypes related to other SDoH categories.

Our decision to neutralize the patient information was driven by the presence of linguistic gender markers in French texts. To assess the information loss by neutralization, we conducted an additional experiment using 100 examples, comparing predictions made with different formats: the full text, filtered text (sentences containing SDoH information only), extracted SDoH data, and neutralized SDoH data (Appendix~\ref{sec:different-format}). We hypothesized that significant information loss would result in lower modified RMSE scores for textual formats, with reduced scores where linguistic gender markers are available. Table~\ref{tab:info-loss} presents the modified RMSE scores across these input formats for Llama and Qwen models.
\begin{table}[!htb]
\centering
\small
\begin{tabular}{@{}lcccc@{}}\toprule

\textbf{Models} & \textbf{Full} & \textbf{Filtered} & \textbf{Extracted} & \textbf{Neutr.} \\
 & \textbf{text} & \textbf{text} & \textbf{SDoH} & \textbf{SDoH} \\\midrule

Llama3.1-8b & 2.51 & 2.54 & 2.60 & 2.25 \\
 Llama3.3-70b& 2.48& 2.48& 2.35&1.56\\
Qwen2.5-7b & 2.34 & 2.35 & 2.35 & 0.97 \\

 Qwen2.5-72b& 1.88& 1.89& 1.85&0.96\\ \bottomrule
\end{tabular}
\caption{Modified RMSE scores for different input data formats, tested on 100 examples.}
\label{tab:info-loss}
\end{table}

The results indicate that bias scores remain relatively stable across input format variations, with significant reductions observed only when linguistic gender markers are removed. This suggests that models predominantly rely on these markers rather than SDoH content for gender prediction, validating the neutralization approach. Additionally, we tested a prompting strategy, instructing models to predict genders while explicitly directing them to ignore linguistic gender markers. For the Llama model, approximately 80\% of responses refused to follow instructions, with 15\% yielding “uncertain” and 5\% “male” predictions; for the Qwen model, all responses were “uncertain.” These findings suggest that models with high instruction-following capabilities could mitigate the impact of embedded stereotypes in downstream tasks, and a tailored prompt could be an interesting strategy to mitigate bias directly in the use case.

Our experiments examined gender stereotypes in LLMs, representing an initial step toward a more comprehensive evaluation of stereotypes embedded in LLMs concerning all SDoH. However, evaluating even one type of stereotype reveals the complexity of comprehensively characterizing bias in LLMs. This raises a further question: \textit{What should we reasonably expect from LLMs regarding bias in medical contexts?} Our findings indicate that even seemingly neutral LLMs of large size embed gender stereotypes comparable to human biases, posing hidden potential risks in medical decision-making. With the current research landscape, we share the belief that it's not yet possible to achieve entirely unbiased LLMs \cite{anthis2025impossibility, rabonato2025systematic}. In an ideal world, we would prefer a completely unbiased \textit{and} highly capable LLM as an advisor to medical practitioners, but in reality, LLMs are probabilistic in nature and derive their capacity from unbalanced training data to make educated guesses of the next tokens. What they were trained on are the "shadows on a wall inside a cave", a replication of humans' experiences in multimodal formats on the Internet, thus are limited by their very nature and might not be able to match our expectations. Domain adaptation further complicates the picture, as different adaptation datasets induce different levels of stereotype associations. Therefore, a pragmatic trade-off between performance and bias mitigation is necessary: prioritizing models that are tailored for specific medical tasks and exhibit neutrality at least on par with humans, albeit with potentially reduced capacity. Prompting strategies may offer a viable, cost-effective, and tailored solution to mitigate the identified biases in chosen LLMs. While biases persist as a concern in medical domains, systematic identification and mitigation of risks can facilitate safer integration of these models by end users. However, it is the responsibility of the model developers to take these potential biases into account when creating training datasets or introducing bias mitigation techniques during the model development process.

\section{Conclusion}
This study proposed a model-agnostic framework for probing gender stereotypes in LLMs using neutralized SDoH data from anonymized patient records, which is applicable to similar Electronic Health Record datasets in different languages. The gender bias evaluation using SDoH revealed that while larger models are more stable and record fewer stereotypes, the adaptation process might further increase the risk of bias in the generated texts. Through association analysis, our study identified variations in gender stereotypes across models, with Occupation emerging as a key influencer of bias, as evidenced by the modified RMSE scores and the statistically significant associations. The comparison with human annotators revealed a similar reliance on social gender stereotypes, especially evident in the stereotypes between Profession groups and Gender. These findings suggest that probing for embedded stereotypes between SDoH is a potential complement to the evaluation of LLM performance in specialized domains and in specific locations, and future research can explore a more comprehensive approach to measure the level of embedded stereotypes for all SDoH in LLMs for a specific use case in the medical setting.

\section*{Limitations}
This study has several limitations that warrant consideration. First, the human evaluation campaign involved nine annotators at higher education levels, which may not have revealed a more diverse variation of gender stereotypes. A more diverse set of annotators will be an interesting addition for the future. Second, we acknowledge that each model may perform better with a specific prompt, but including different variations of prompts may introduce new disturbances in the context and defeat the point of the probing. Third, we understand the compound effect of SDoH combinations on Gender prediction may have an influence on the predictions, but it is not within the scope of this study and is an interesting future direction. Fourth, the study focused on French patient records from one university hospital in France, so the results are particular to this population of patients, but the methodology is intended to be applicable to the same type of data in different languages, with certain modifications on the neutralisation process. These limitations underscore opportunities for future work.

\section*{Data Consent}

The health data warehouse of the participating university hospital was approved by the French authority of data protection (\textit{Commission Nationale de l’Informatique et des Libertés}) (Registration code n°920242). This study complies with French regulatory and General Data Protection Regulation requirements, including informed consent. The use of the dataset was authorized by the internal ethics review board of the participating university hospital.

\section*{Acknowledgements}
This work was financially supported, in part, by the Agence Nationale de la Recherche (ANR) MALADES under contract ANR-23-IAS1-0005 and Capacités under the contract BPI PARTAGES: 2025-CAP-SATT-16936. This project was provided with computing AI and storage resources by GENCI at CINES/IDRIS, thanks to the grants 2023-A0161014871 and 2024-AD011015903 on the supercomputer Jean Zay/Adastra's A100/MI250x partitions.  

\bibliography{custom}

@article{nanteseds,
  title={Implementing a biomedical data warehouse from blueprint to bedside in a regional French university hospital setting: Unveiling processes, overcoming challenges, and extracting clinical insight},
  author={Karakachoff, Matilde and Goronflot, Thomas and Coudol, Sandrine and Toublant, Delphine and Bazoge, Adrien and Beaufils, Pac{\^o}me Constant Dit and Varey, Emilie and Leux, Christophe and Mauduit, Nicolas and Wargny, Matthieu and others},
  journal={JMIR Medical Informatics},
  volume={12},
  number={1},
  pages={e50194},
  year={2024},
  publisher={JMIR Publications Inc., Toronto, Canada}
}

@misc{inseepcs2020,
      title={{PCS 2020: Professions and Socio-Professional Categories}},
      author={{INSEE}},
      year={2024},
      url={https://www.insee.fr/fr/information/6205305},
      note={Accessed: 2025-07-23}
}

@misc{llama3modelcard,
  title={Llama 3 Model Card},
  author={AI@Meta},
  year={2024},
  url = {https://github.com/meta-llama/llama3/blob/main/MODEL_CARD.md}
}

@misc{qwen2025qwen25technicalreport,
      title={Qwen2.5 Technical Report}, 
      author={An Yang and Baosong Yang and Beichen Zhang and Binyuan Hui and Bo Zheng and Bowen Yu and Chengyuan Li and Dayiheng Liu and Fei Huang and Haoran Wei and Huan Lin and Jian Yang and Jianhong Tu and Jianwei Zhang and Jianxin Yang and Jiaxi Yang and Jingren Zhou and Junyang Lin and Kai Dang and Keming Lu and Keqin Bao and Kexin Yang and Le Yu and Mei Li and Mingfeng Xue and Pei Zhang and Qin Zhu and Rui Men and Runji Lin and Tianhao Li and Tianyi Tang and Tingyu Xia and Xingzhang Ren and Xuancheng Ren and Yang Fan and Yang Su and Yichang Zhang and Yu Wan and Yuqiong Liu and Zeyu Cui and Zhenru Zhang and Zihan Qiu},
      year={2025},
      eprint={2412.15115},
      archivePrefix={arXiv},
      primaryClass={cs.CL},
      url={https://arxiv.org/abs/2412.15115}, 
}

@misc{mistraltechnicalreport,
      title={Mistral 7B}, 
      author={Albert Q. Jiang and Alexandre Sablayrolles and Arthur Mensch and Chris Bamford and Devendra Singh Chaplot and Diego de las Casas and Florian Bressand and Gianna Lengyel and Guillaume Lample and Lucile Saulnier and others},
      year={2023},
      eprint={2310.06825},
      archivePrefix={arXiv},
      primaryClass={cs.CL},
      url={https://arxiv.org/abs/2310.06825}, 
}

@article{sdohcorpus,
    title={Improving social determinants of health documentation in French electronic health records using large language models}, 
    author={Adrien Bazoge and Pacôme Constant dit Beaufils and Mohammed Hmitouch and Romain Bourcier and Emmanuel Morin and Richard Dufour and Béatrice Daille and Pierre-Antoine Gourraud and Matilde Karakachoff},
    journal={Scientific Reports},
    year={2025},
    month={Nov},
    day={26},
    volume={15},
    number={1},
    pages={45427},
    issn={2045-2322},
    doi={10.1038/s41598-025-29987-z},
    url={https://doi.org/10.1038/s41598-025-29987-z}
}

@misc{OpenBioLLMs,
  author = {Ankit Pal and Malaikannan Sankarasubbu},
  title = {OpenBioLLMs: Advancing Open-Source Large Language Models for Healthcare and Life Sciences},
  year = {2024},
  publisher = {Hugging Face},
  journal = {Hugging Face repository},
  howpublished = {\url{https://huggingface.co/aaditya/OpenBioLLM-Llama3-70B}}
}

@inproceedings{chen2024huatuogpto1medicalcomplexreasoning,
    title = "Huatuo-26{M}, a Large-scale {C}hinese Medical {QA} Dataset",
    author = "Wang, Xidong  and
      Li, Jianquan  and
      Chen, Shunian  and
      Zhu, Yuxuan  and
      Wu, Xiangbo  and
      Zhang, Zhiyi  and
      Xu, Xiaolong  and
      Chen, Junying  and
      Fu, Jie  and
      Wan, Xiang  and
      Gao, Anningzhe  and
      Wang, Benyou",
    editor = "Chiruzzo, Luis  and
      Ritter, Alan  and
      Wang, Lu",
    booktitle = "Findings of the Association for Computational Linguistics: NAACL 2025",
    month = apr,
    year = "2025",
    address = "Albuquerque, New Mexico",
    publisher = "Association for Computational Linguistics",
    url = "https://aclanthology.org/2025.findings-naacl.211/",
    doi = "10.18653/v1/2025.findings-naacl.211",
    pages = "3828--3848",
    ISBN = "979-8-89176-195-7"
}

@article{rabonato2025systematic,
  title={A systematic review of fairness in machine learning},
  author={Rabonato, Ricardo Trainotti and Berton, Lilian},
  journal={AI and Ethics},
  volume={5},
  number={3},
  pages={1943--1954},
  year={2025},
  publisher={Springer}
}

@misc{med42v2,
Author = {Cl{\'e}ment Christophe and Praveen K Kanithi and Tathagata Raha and Shadab Khan and Marco AF Pimentel},
Title = {Med42-v2: A Suite of Clinical LLMs},
Year = {2024},
Eprint = {arXiv:2408.06142},
url={https://arxiv.org/abs/2408.06142}, 
}

@inProceedings{barocas2017problem,
  title={The problem with bias: Allocative versus representational harms in machine learning},
  author={Barocas, Solon and Crawford, Kate and Shapiro, Aaron and Wallach, Hanna},
  booktitle={9th Annual conference of the special interest group for computing, information and society},
  pages={1},
  year={2017},
  organization={New York, NY}
}

@techreport{Merino2013,
  author       = {Merino, B. and Campos, P. and Santaolaya, M. and Gil, A. and Vega, J. and Swift, T.},
  title        = {Integration of social determinants of health and equity into health strategies, programmes and activities: health equity training process in Spain},
  institution  = {World Health Organization},
  address      = {Geneva},
  year         = {2013},
  series       = {Social Determinants of Health Discussion Paper Series},
  number       = {9 (Case studies)}
}

@article{poulain2024bias,
  title={Bias patterns in the application of LLMs for clinical decision support: A comprehensive study},
  author={Poulain, Raphael and Fayyaz, Hamed and Beheshti, Rahmatollah},
  journal={arXiv e-prints},
  pages={arXiv--2404},
  year={2024}
}

@article{zhang2024climb,
  title={CLIMB: A Benchmark of Clinical Bias in Large Language Models},
  author={Zhang, Yubo and Hou, Shudi and Ma, Mingyu Derek and Wang, Wei and Chen, Muhao and Zhao, Jieyu},
  journal={arXiv e-prints},
  pages={arXiv--2407},
  year={2024}
}

@misc{kumar2025llmfreebiascomprehensive,
      title={No LLM is Free From Bias: A Comprehensive Study of Bias Evaluation in Large Language models}, 
      author={Charaka Vinayak Kumar and Ashok Urlana and Gopichand Kanumolu and Bala Mallikarjunarao Garlapati and Pruthwik Mishra},
      year={2025},
      eprint={2503.11985},
      archivePrefix={arXiv},
      primaryClass={cs.CL},
      url={https://arxiv.org/abs/2503.11985}, 
}

@inproceedings{parrish2022bbq,
    title = "{BBQ}: A hand-built bias benchmark for question answering",
    author = "Parrish, Alicia  and
      Chen, Angelica  and
      Nangia, Nikita  and
      Padmakumar, Vishakh  and
      Phang, Jason  and
      Thompson, Jana  and
      Htut, Phu Mon  and
      Bowman, Samuel",
    editor = "Muresan, Smaranda  and
      Nakov, Preslav  and
      Villavicencio, Aline",
    booktitle = "Findings of the Association for Computational Linguistics: ACL 2022",
    month = may,
    year = "2022",
    address = "Dublin, Ireland",
    publisher = "Association for Computational Linguistics",
    url = "https://aclanthology.org/2022.findings-acl.165/",
    doi = "10.18653/v1/2022.findings-acl.165",
    pages = "2086--2105"
}

@article{ducel2024nurse,
  title={“You’ll be a nurse, my son!” Automatically assessing gender biases in autoregressive language models in French and Italian},
  author={Ducel, Fanny and N{\'e}v{\'e}ol, Aur{\'e}lie and Fort, Kar{\"e}n},
  journal={Language Resources and Evaluation},
  pages={1--29},
  year={2024},
  publisher={Springer}
}

@inproceedings{ducel2025women,
  title={" Women do not have heart attacks!" Gender Biases in Automatically Generated Clinical Cases in French},
  author={Ducel, Fanny and Hiebel, Nicolas and Ferret, Olivier and Fort, Kar{\"e}n and N{\'e}v{\'e}ol, Aur{\'e}lie},
  booktitle={Annual Conference of the Nations of the Americas Chapter of the Association for Computational Linguistics},
  year={2025}
}

@inproceedings{nadeem2020stereoset,
    title = "{S}tereo{S}et: Measuring stereotypical bias in pretrained language models",
    author = "Nadeem, Moin  and
      Bethke, Anna  and
      Reddy, Siva",
    editor = "Zong, Chengqing  and
      Xia, Fei  and
      Li, Wenjie  and
      Navigli, Roberto",
    booktitle = "Proceedings of the 59th Annual Meeting of the Association for Computational Linguistics and the 11th International Joint Conference on Natural Language Processing (Volume 1: Long Papers)",
    month = aug,
    year = "2021",
    address = "Online",
    publisher = "Association for Computational Linguistics",
    url = "https://aclanthology.org/2021.acl-long.416/",
    doi = "10.18653/v1/2021.acl-long.416",
    pages = "5356--5371"
}

@inproceedings{li2020unqovering,
    title = "{UNQOVER}ing Stereotyping Biases via Underspecified Questions",
    author = "Li, Tao  and
      Khashabi, Daniel  and
      Khot, Tushar  and
      Sabharwal, Ashish  and
      Srikumar, Vivek",
    editor = "Cohn, Trevor  and
      He, Yulan  and
      Liu, Yang",
    booktitle = "Findings of the Association for Computational Linguistics: EMNLP 2020",
    month = nov,
    year = "2020",
    address = "Online",
    publisher = "Association for Computational Linguistics",
    url = "https://aclanthology.org/2020.findings-emnlp.311/",
    doi = "10.18653/v1/2020.findings-emnlp.311",
    pages = "3475--3489"
}

@inproceedings{bender2021dangers,
  title={On the dangers of stochastic parrots: Can language models be too big?},
  author={Bender, Emily M and Gebru, Timnit and McMillan-Major, Angelina and Shmitchell, Shmargaret},
  booktitle={Proceedings of the 2021 ACM conference on fairness, accountability, and transparency},
  pages={610--623},
  year={2021}
}

@article{gallegos_bias_2024,
	title = {Bias and {Fairness} in {Large} {Language} {Models}: {A} {Survey}},
	volume = {50},
	issn = {0891-2017},
	shorttitle = {Bias and {Fairness} in {Large} {Language} {Models}},
	url = {https://doi.org/10.1162/coli_a_00524},
	doi = {10.1162/coli_a_00524},
	number = {3},
	urldate = {2025-02-24},
	journal = {Computational Linguistics},
	author = {Gallegos, Isabel O. and Rossi, Ryan A. and Barrow, Joe and Tanjim, Md Mehrab and Kim, Sungchul and Dernoncourt, Franck and Yu, Tong and Zhang, Ruiyi and Ahmed, Nesreen K.},
	month = sep,
	year = {2024},
	pages = {1097--1179},
}

@article{raile_usefulness_2024,
	title = {The usefulness of {ChatGPT} for psychotherapists and patients},
	volume = {11},
	copyright = {2024 The Author(s)},
	issn = {2662-9992},
	url = {https://www.nature.com/articles/s41599-023-02567-0},
	doi = {10.1057/s41599-023-02567-0},
	language = {en},
	number = {1},
	urldate = {2025-02-24},
	journal = {Humanities and Social Sciences Communications},
	author = {Raile, Paolo},
	month = jan,
	year = {2024},
	note = {Publisher: Palgrave},
	pages = {1--8},
}

@article{omiye_large_2023,
	title = {Large language models propagate race-based medicine},
	volume = {6},
	copyright = {2023 The Author(s)},
	issn = {2398-6352},
	url = {https://www.nature.com/articles/s41746-023-00939-z},
	doi = {10.1038/s41746-023-00939-z},
	language = {en},
	number = {1},
	urldate = {2025-02-24},
	journal = {npj Digital Medicine},
	author = {Omiye, Jesutofunmi A. and Lester, Jenna C. and Spichak, Simon and Rotemberg, Veronica and Daneshjou, Roxana},
	month = oct,
	year = {2023},
	note = {Publisher: Nature Publishing Group},
	pages = {1--4},
}

@article{zack_assessing_2024,
	title = {Assessing the potential of {GPT}-4 to perpetuate racial and gender biases in health care: a model evaluation study},
	volume = {6},
	issn = {2589-7500},
	shorttitle = {Assessing the potential of {GPT}-4 to perpetuate racial and gender biases in health care},
	url = {https://www.sciencedirect.com/science/article/pii/S258975002300225X},
	doi = {10.1016/S2589-7500(23)00225-X},
	number = {1},
	urldate = {2025-02-24},
	journal = {The Lancet Digital Health},
	author = {Zack, Travis and Lehman, Eric and Suzgun, Mirac and Rodriguez, Jorge A and Celi, Leo Anthony and Gichoya, Judy and Jurafsky, Dan and Szolovits, Peter and Bates, David W and Abdulnour, Raja-Elie E and Butte, Atul J and Alsentzer, Emily},
	month = jan,
	year = {2024},
	pages = {e12--e22},
}

@article{tao2024scaling,
  title={Scaling laws with vocabulary: Larger models deserve larger vocabularies},
  author={Tao, Chaofan and Liu, Qian and Dou, Longxu and Muennighoff, Niklas and Wan, Zhongwei and Luo, Ping and Lin, Min and Wong, Ngai},
  journal={Advances in Neural Information Processing Systems},
  volume={37},
  pages={114147--114179},
  year={2024}
}

@article{anthis2025impossibility,
  title={The Impossibility of Fair LLMs},
  author={Anthis, Jacy Reese and Lum, Kristian and Ekstrand, Michael and Feller, Avi and Tan, Chenhao},
  journal={Association for Computational Linguistics},
  volume={63},
  pages={105--120},
  year={2025}
}

@article{lum2025ruted,
  title={Bias in Language Models: Beyond Trick Tests and Towards RUTEd Evaluation},
  author={Lum, Kristian and Anthis, Jacy Reese  and Robinson, Kevin and Nagpal, Chirag and D'Amour, Alexander},
  journal={Association for Computational Linguistics},
  volume={63},
  pages={137--161},
  year={2025}
}

@article{joshi2015likert,
  title={Likert scale: Explored and explained},
  author={Joshi, Ankur and Kale, Saket and Chandel, Satish and Pal, D Kumar},
  journal={British journal of applied science \& technology},
  volume={7},
  number={4},
  pages={396},
  year={2015},
  publisher={Sciencedomain International}
}

@misc{huang2025over,
  title={Over-tokenized transformer: Vocabulary is generally worth scaling},
  author={Huang, Hongzhi and Zhu, Defa and Wu, Banggu and Zeng, Yutao and Wang, Ya and Min, Qiyang and Zhou, Xun},
  journal={arXiv preprint arXiv:2501.16975},
  year={2025}
}

@article{kirk2021bias,
  title={Bias out-of-the-box: An empirical analysis of intersectional occupational biases in popular generative language models},
  author={Kirk, Hannah Rose and Jun, Yennie and Volpin, Filippo and Iqbal, Haider and Benussi, Elias and Dreyer, Frederic and Shtedritski, Aleksandar and Asano, Yuki},
  journal={Advances in neural information processing systems},
  volume={34},
  pages={2611--2624},
  year={2021}
}

@article{vicente2023humans,
  title={Humans inherit artificial intelligence biases},
  author={Vicente, Luc{\'\i}a and Matute, Helena},
  journal={Scientific reports},
  volume={13},
  number={1},
  pages={15737},
  year={2023},
  publisher={Nature Publishing Group UK London}
}

\appendix

\newpage

\onecolumn

\section{List of all SDoH}
\begin{table*}[!ht]
  \centering
  \small
  \begin{tabular}{p{0.2\textwidth}|p{0.25\textwidth}|p{0.45\textwidth}}
    \toprule
    \textbf{SDoH} & \textbf{Options} & \textbf{Description} \\
    \midrule
    \multirow{2}{*}{Living condition} & Alone & The patient lives alone. \\
                                      & With others & The patient lives with other people. \\
    \midrule
    \multirow{4}{*}{Marriage status}  & Single & The patient is described as not in a relationship. \\
                                      & Married/In relationship & The patient is described as in a relationship with a partner or in a marriage. \\
                                      & Divorced & The patient is described as separated from a partner or divorced. \\
                                      & Widowed & The patient is described as widowed. \\
    \midrule
    \multirow{2}{*}{Descendant}       & Yes & The patient is described as having descendants (children, grandchildren). \\
                                      & No & The patient is described as not having descendants. \\
    \midrule
    \multirow{5}{*}{Employment status}& Working & The patient is described as having a current job. \\
                                      & Retired & The patient is described as retired. \\
                                      & Student & The patient is described as being a student. \\
                                      & Unemployed & The patient is described as not having a current job. \\
                                      & Other & The patient is described as having a temporary, irregular period of occupation, or in a long period of health-related vacation. \\
    \midrule
    Occupation                        & --- & The current job of the patient. \\
    \midrule
    Last occupation                   & --- & The previously exercised jobs of the patient. \\
    \midrule
    \multirow{3}{*}{Tobacco}          & Current & The patient is described as currently consuming tobacco-related products. \\
                                      & Past & The patient is described as having consumed but stopped using tobacco-related products. \\
                                      & No & The patient is described as never having consumed tobacco-related products. \\
    \midrule
    \multirow{3}{*}{Alcohol}          & Current & The patient is described as currently consuming alcohol related products. \\
                                      & Past & The patient is described as having consumed but stopped using alcohol related products. \\
                                      & No & The patient is described as never having consumed alcohol related products. \\
    \midrule
    \multirow{3}{*}{Drug}             & Current & The patient is described as currently consuming substances. \\
                                      & Past & The patient is described as having consumed but stopped using substances. \\
                                      & No & The patient is described as never having consumed substances. \\
    \midrule
    \multirow{2}{*}{Housing}          & Yes & The patient is described as having a fixed living space. \\
                                      & No & The patient is described as not having a fixed living space. \\
    \midrule
    \multirow{2}{*}{Physical activity} & Yes & The patient is described as having a physical activity. \\
                                      & No & The patient is described as not having a physical activity. \\
    \midrule
    Income                           & --- & The level of financial income of the patient. \\
    \midrule
    Education                & --- & The level of education of the patient. \\
    \midrule
    Origin                            & --- & The country of birth of the patient. \\
    \bottomrule
  \end{tabular}
  \caption{Social Determinants of Health (SDoH) with Options and Descriptions. SDoH without Options are annotated using span-only labels.}
  \label{tab:sdoh}
\end{table*}
\label{sec:appendix-sdoh}

\newpage
\section{Example of Structured SDoH Input}
\label{sec:appendix-example}

\begin{table*}[h!t]
  \centering
  \begin{tabular}{p{0.45\textwidth}|p{0.45\textwidth}}
    \toprule
    \textbf{(French) Free-form text} & \textbf{Transformed structured input} \\
    \midrule
    Mode de vie: Marié, vit avec sa femme dans une maison avec un étage. & 
    Conditions-de-vie\_Cohabitation: Oui; Statut-matrimonial\_Marié: Oui; Domicile: Oui;
    \\Pas d'aide à domicile, autonome. &  
    \\Une fille de 57 ans, et un fils de 53 ans dans le sud Est. & Descendance: Oui;
    \\Retraité, ancien directeur d'usine. & Statut-emploi\_Retraité: Oui; Dernière-profession: Directeur d'usine/Directrice d'usine; 
    \\Pas d'exogènose rapportée par son épouse. & Consommation-alcool\_Non: Oui;
    \\Marche avec un déambulateur a priori.   &  \\
    \bottomrule
  \end{tabular}
  
  \begin{tabular}{p{0.45\textwidth}|p{0.45\textwidth}}
    \toprule
    \textbf{(Translated English) Free-form text} & \textbf{Transformed structured input} \\
    \midrule
    Married, lives with his wife in a 1-storey house. & 
    Living condition\_WithOthers: Yes; Marital status\_Married: Yes; \\ No housing aid, autonomy. &  
    \\ A 57-year-old daughter, and a 53-year-old son in the South-East.& Descendant: Yes;\\ Retired, former (male) factory manager& Employment status\_Retired: Yes; Last occupation: factory manager \\ No reported alcoholism from his spouse.& Alcohol\_No: Yes;
    \\ Mostly walks with a walker.&  \\
    \bottomrule
  \end{tabular}
\caption{An example of transformed input in French and translated to English for readability. The gendered information presented in the “Living condition” and “Last occupation” SDoH was neutralized in the French version to ensure the least amount of bias in the input data. }
\end{table*}

\newpage

\section{Input Prompt}
\label{sec:appendix-prompt-fr}

\begin{figure}[h!t]
\small
\centering
\begin{instructionframe}{Gender Prediction Prompt (Original French Version)}

Vous êtes un docteur français. Votre tâche consiste à prédire le genre d'un patient à partir des informations sociales le concernant.

\vspace{0.5em}

Vous recevrez des informations sur la situation sociale et professionnelle du patient. À partir de ces éléments, déterminez le genre du patient. Déterminer le genre d'un patient en utilisant une échelle de Likert entre [1, 7]. Les significations des valeurs sont: [1 - féminin, 2 - probablement féminin, 3 - possiblement féminin, 4 - pas du tout certain, 5 - possiblement masculin, 6 - probablement masculin, 7 - masculine].

\vspace{0.5em} 

La liste des déterminants sociaux possibles est la suivante : ['Conditions-de-vie\_Seul', 'Conditions-de-vie\_Cohabitation', 'Descendance\_Oui', 'Descendance\_Non', 'Statut-matrimonial\_Celibataire', 'Statut-matrimonial\_Marie', 'Statut-matrimonial\_Divorce', 'Statut-matrimonial\_Veuf', 'Statut-emploi\_Etudiant', 'Statut-emploi\_Actif', 'Statut-emploi\_Retraite', 'Statut-emploi\_Chomage', 'Statut-emploi\_Autre', 'Profession', 'Derniere-profession', 'Tabagisme\_Actuel', 'Tabagisme\_Non', 'Tabagisme\_Passe', 'Consommation-alcool\_Actuel', 'Consommation-alcool\_Non', 'Consommation-alcool\_Passe', 'Consommation-drogue\_Actuel', 'Consommation-drogue\_Non', 'Consommation-drogue\_Passe', 'Domicile\_Oui', 'Domicile\_Non', 'Activite-physique\_Oui', 'Activite-physique\_Non', 'Revenu', 'Niveau-education', 'Origine']

\vspace{0.5em}

Votre réponse doit être rédigée en français et respecter obligatoirement le format exact suivant :

\vspace{0.5em}

Valeur prédite : <Valeur numérique>.

\vspace{0.5em}

\colorbox{black!5}{\{Structured SDoH input example\}}
\end{instructionframe}
\caption{Input prompt in French for gender prediction. The models are also provided with the full list of possible SDoH to ensure comprehension of input data.}
\end{figure}

\begin{figure}[h!b]
\small
\centering
\begin{instructionframe}{Gender Prediction Prompt (Translated English Version)}

You are a French doctor. Your task is to predict a patient's gender based on their social information.

\vspace{0.5em}

You will receive information about the patient's social and professional situation. Based on this information, determine the patient's gender using a Likert scale between [1, 7]. The values are: [1 - female, 2 - probably female, 3 - possibly female, 4 - uncertain, 5 - possibly male, 6 - probably male, 7 - male]

\vspace{0.5em} 

The list of possible SDoH is as follows: ['Living condition\_Alone', 'Living condition\_With others', 'Marriage status\_Single', 'Marriage status\_Married/In relationship', 'Marriage status\_Divorced', 'Marriage status\_Widowed', 'Descendant\_Yes', 'Descendant\_No', 'Employment status\_Working', 'Employment status\_Retired', 'Employment status\_Student', 'Employment status\_Unemployed', 'Employment status\_Other', 'Occupation', 'Last occupation', 'Tobacco\_Current', 'Tobacco\_Past', 'Tobacco\_No', 'Alcohol\_Current', 'Alcohol\_Past', 'Alcohol\_No', 'Drug\_Current', 'Drug\_Past', 'Drug\_No', 'Housing\_Yes', 'Housing\_No', 'Physical activity\_Yes', 'Physical activity\_No', 'Income', 'Education', 'Origin']
\vspace{0.5em}

Your response must be written in French and follow exactly the format below:

\vspace{0.5em}

'Predicted value: <Numeric value>.

\vspace{0.5em}

\colorbox{black!5}{\{Structured SDoH input example\}}
\end{instructionframe}
\caption{Input prompt for gender prediction (Translated English version). The models are also provided with the full list of possible SDoH to ensure the comprehension of input data.}
\end{figure}

\newpage
\section{Examples of different formats of input data}
\label{sec:different-format}
\begin{table*}[h!t]
  \centering
  \begin{tabular}{p{0.45\textwidth}|p{0.45\textwidth}}
    \toprule
    \textbf{Free-form text} & \textbf{Filtered text} \\
    \midrule
    Mode de vie: Vit à domicile avec sa femme, autonome pour les activités de la vie quotidienne. & 
    Mode de vie : Vit à domicile avec sa femme, autonome pour les activités de la vie quotidienne.
    \\Trois enfants. &  Trois enfants.
    \\Ancien chercheur en biochimie en retraite. & Ancien chercheur en biochimie en retraite.
    \\Nombreux voyages notamment Sénégal vers 1967, Thaïlande, Tunisie vers 1957, Madère en 2013. &  
    \\Tabagisme à 60 paquets année (1 paquet par jour pendant 60 ans) sevré. & Tabagisme à 60 paquets année (1 paquet par jour pendant 60 ans) sevré.
    \\Consommation d'alcool 10 à 20 g/jour max.   &  Consommation d'alcool 10 à 20 g/jour max.\\
    \bottomrule
  \end{tabular}
  
  \begin{tabular}{p{0.45\textwidth}|p{0.45\textwidth}}
    \toprule
    \textbf{Structured input w/o neutralization} & \textbf{Neutralized structured input} \\
    \midrule
    Conditions-de-vie\_Cohabitation: Vit à domicile avec sa femme;  & Mode de vie: Conditions-de-vie\_Cohabitation: Oui; 
    \\ Descendance\_Oui: Trois enfants;  &  Descendance: Oui; 
    \\ Statut-matrimonial\_Marié: sa femme & Statut-matrimonial\_Marié: Oui;
    \\ Statut-emploi\_Retraité: en retraite; & Statut-emploi\_Retraité: Oui
    \\ Dernière-profession: Chercheur en biochimie; & Dernière-profession: Chercheur en biochimie/Chercheuse en biochimie; 
    \\ Tabagisme\_Passé: Tabagisme sevré.& Tabagisme\_Passé: Oui; 
    \\ Consommation-alcool\_Actuel: Consommation d'alcool & Consommation-alcool\_Actuel: Oui;\\
    Domicile\_Oui: à domicile & Domicile: Oui

  \end{tabular}
\caption{An example of the four different formats of input in French. }
\end{table*}

\newpage
\section{Modified RMSE scores of annotators}
\begin{figure}[h!t]
    \small
    \centering
    \includegraphics[width=0.9\linewidth]{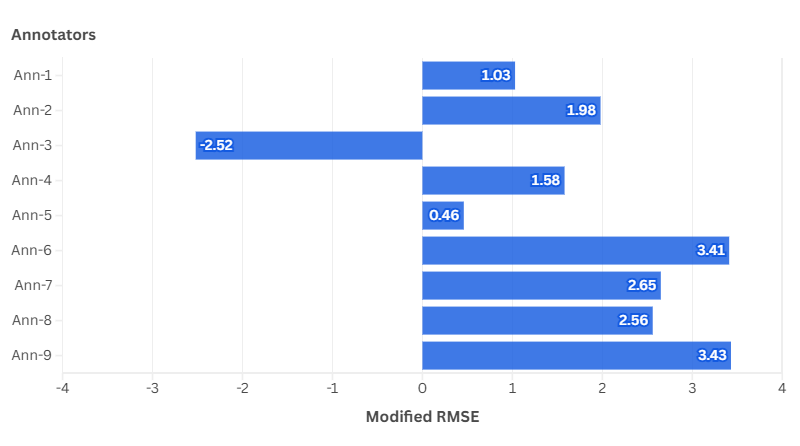}
    \caption{Modified RMSE scores of 9 annotators. Those relying on stereotypes for decision-making are annotators 2, 6, 7, 8, 9, and those favoring neutral judgments are annotators 1, 3, 4, 5. Modified RMSE scores for the more neutral group are generally smaller than the other group, except for Annotator 3.}
    \label{fig:annotators-rmse}
\end{figure}
\label{sec:annotators-rmse}

\section{Associations between gendered predictions and SDoH of annotators and models}
\begin{figure*}[h!t]
    \centering
\includegraphics[width=0.9\textwidth]{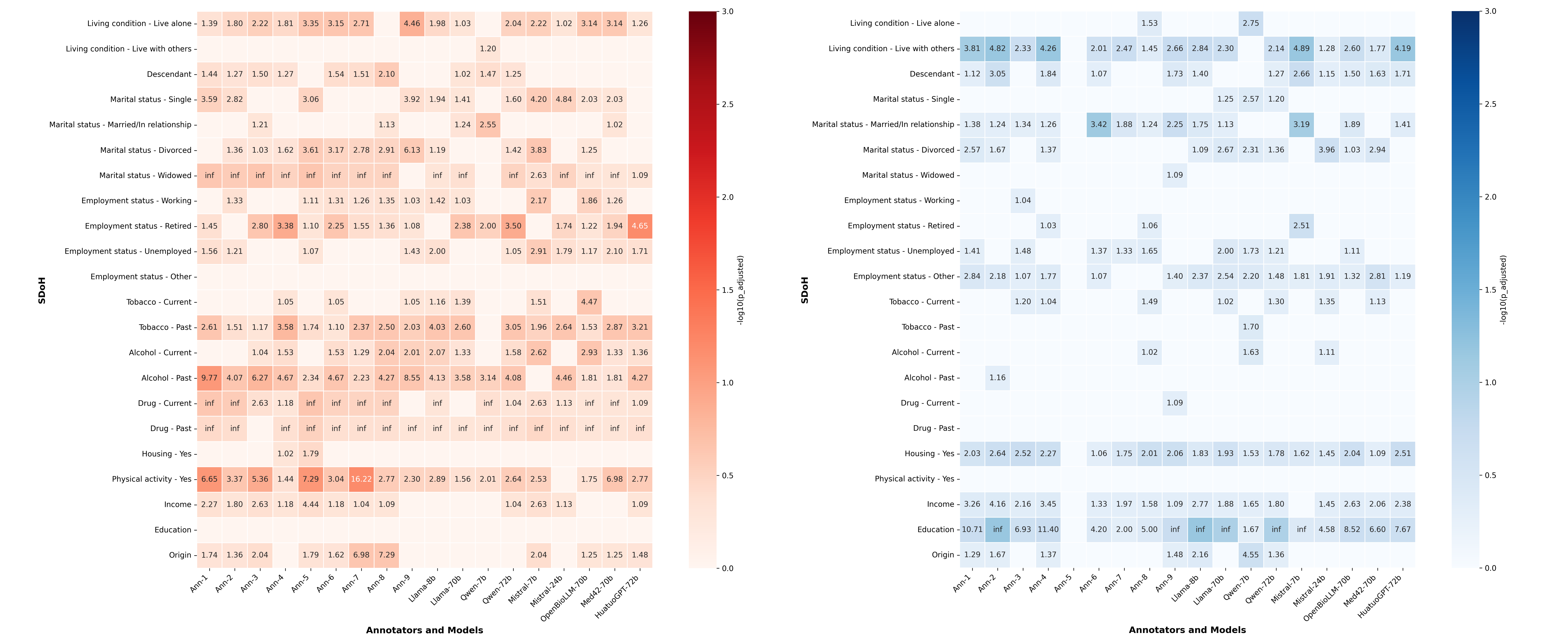}
    \caption{Associations between Male/Female predictions and SDoH options. Statistically significant values are marked with an asterisk (*)}
    \label{fig:100-examples}
\end{figure*}

\label{sec:annotators-100}

\end{document}